\documentclass{ecai} 
\usepackage{latexsym}
\usepackage{amssymb}
\usepackage{amsmath}
\usepackage{amsthm}
\usepackage{booktabs}
\usepackage{enumitem}
\usepackage{graphicx}
\usepackage{color}
\usepackage{float}
\usepackage{booktabs}
\usepackage{colortbl}

\newcommand{\BibTeX}{B\kern-.05em{\sc i\kern-.025em b}\kern-.08em\TeX}

\DeclareRobustCommand\onedot{\futurelet\@let@token\@onedot}

\def\ie{\emph{i.e., }}    
   
\def\etc{\emph{etc. }}

\begin{document}

\begin{frontmatter}

\title{CorrAdaptor: Adaptive Local Context Learning for Correspondence Pruning}

\author[A]{Wei Zhu}
\author[B]{\fnms{Yicheng}~\snm{Liu}}
\author[B]{\fnms{Yuping}~\snm{He}}
\author[B]{\fnms{Tangfei}~\snm{Liao}}
\author[A]{\fnms{Kang}~\snm{Zheng}}
\author[A]{\fnms{Xiaoqiu}~\snm{Xu}}
\author[B]{\fnms{Tao}~\snm{Wang}\thanks{Corresponding Authors. Email: taowangzj@gmail.com, lutong@nju.edu.cn.}}
\author[B]{\fnms{Tong}~\snm{Lu}\footnotemark[*]}
\address[A]{China Mobile Zijin Innovation Insititute}
\address[B]{Nanjing University}

\begin{abstract}
In the fields of computer vision and robotics, accurate pixel-level correspondences are essential for enabling advanced tasks such as structure-from-motion and simultaneous localization and mapping. Recent correspondence pruning methods usually focus on learning local consistency through k-nearest neighbors, which makes it difficult to capture robust context for each correspondence. We propose CorrAdaptor, a novel architecture that introduces a dual-branch structure capable of adaptively adjusting local contexts through both explicit and implicit local graph learning. Specifically, the explicit branch uses KNN-based graphs tailored for initial neighborhood identification, while the implicit branch leverages a learnable matrix to softly assign neighbors and adaptively expand the local context scope, significantly enhancing the model's robustness and adaptability to complex image variations. Moreover, we design a motion injection module to integrate motion consistency into the network to suppress the impact of outliers and refine local context learning, resulting in substantial performance improvements. The experimental results on extensive correspondence-based tasks indicate that our CorrAdaptor achieves state-of-the-art performance both qualitatively and quantitatively. 
The code and pre-trained models are available at \url{https://github.com/TaoWangzj/CorrAdaptor}.

\end{abstract}

\end{frontmatter}
\section{Introduction}
\label{sec:intro}

In the fields of computer vision and robotics, high-quality pixel-level correspondences are fundamental to performing a variety of key tasks, including structure from motion \cite{schonberger2016structure}, simultaneous localization and mapping \cite{mur2015orb}, visual localization \cite{sattler2018benchmarking}, and image stitching \cite{ma2019infrared}. Typically, these correspondences are derived from off-the-shelf detector-descriptors. However, due to complex variations between images, such as rotations, scale, viewpoint, and lighting changes, the initial correspondences often contain a high proportion of mismatches, \ie outliers. Thus, correspondence pruning is essential for image matching and has attracted the attention of many researchers. 

Traditional methods like RANSAC~\cite{fischler1981random} and its variants \cite{barath2019magsac, chum2005matching, raguram2012usac, torr2000mlesac} typically employ iterative sampling strategies to identify correct correspondences. However, their running time significantly increases in scenarios with a high number of outliers, making them less efficient. 
\begin{figure}[t]
    \centering
    \includegraphics[width=0.45\textwidth]{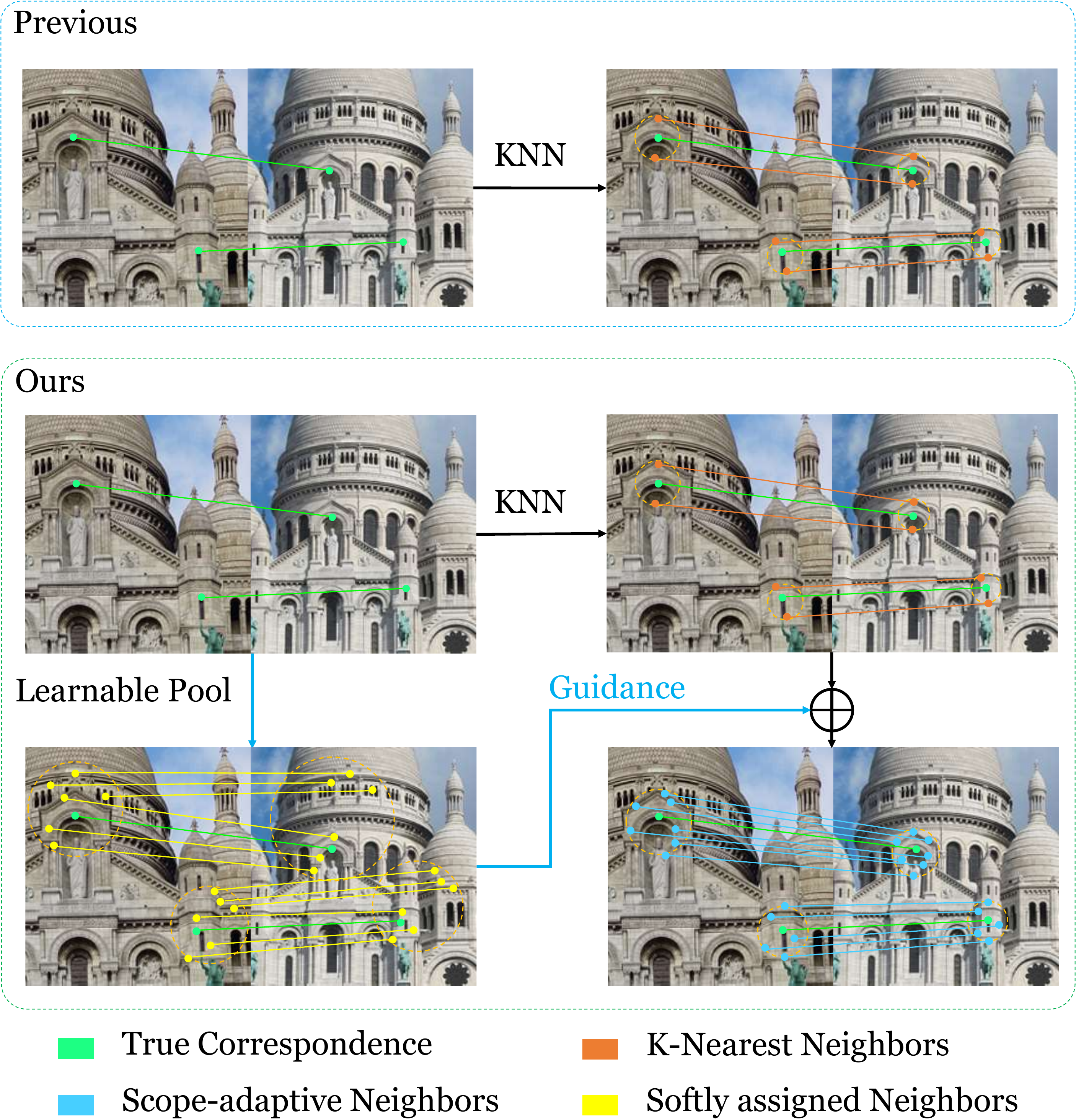}
    \caption{Comparison between the previous pipeline and CorrAdaptor. \textbf{Top}: the previous pipeline focuses on $k$-nearest neighbors for local context learning. \textbf{Bottom}: our CorrAdaptor introduces a dual-branch structure that adaptively adjusts the scope of local contexts. Simple out, we begin by generating initial local contexts using k-nearest neighbors, and then inject softly assigned local contexts, which enables our method to capture the scope-adaptive local context for each correspondence.}
    \label{fig:1}
    
\end{figure}
Recently, learning-based approaches such as deep learning have proven to be effective in selecting correspondences. These methods typically treat the task as a classification problem, using multi-layer perceptrons to distinguish between inliers and outliers. Early studies \cite{yi2018learning, zhang2019learning, zhong2021t}  primarily focus on learning global consistency, \ie the global context, to locate inliers and estimate the relative camera poses. Subsequently, more studies \cite{ dai2022ms2dg, dai2024mgnet,liu2023progressive, liu2021learnable, zhao2019nm, zhao2021progressive} highlight the significance of local context for correspondence pruning and propose various methods to enhance network performance. Until recently, most methods \cite{dai2022ms2dg, liu2023progressive, liu2021learnable, zhao2021progressive} explicitly acquire local context, particularly using the KNN algorithm to identify a fixed number of neighbors, as shown in Fig.~\ref{fig:1} top. However, inherently, the number of neighbors for each correspondence in initial matches remains undefined. On the one hand, some methods \cite{dai2022ms2dg, dai2024mgnet,zhao2019nm} experimentally set the number of neighbors to obtain better performance, but this approach lacks robustness. On the other hand, there are some methods \cite{li2023u, zhang2019learning} that employ learnable approaches to capture local context. Nevertheless, these methods often construct coarse-grained local graphs, resulting in an overly broad range of local contexts and inconsistent correspondences, hindering the learning process. 

To address the above problems, we propose a more robust neighborhood learning method called CorrAdaptor, which can adaptively capture the local context for each correspondence. The pipeline of CorrAdaptor is shown in Fig.~\ref{fig:1} bottom. Our CorrAdaptor features a dual-branch structure that adaptively adjusts the local context range for each correspondence. Specifically, we design two branches: an explicit branch and an implicit branch. The former branch uses KNN to search for a small and fixed number of neighbors for each correspondence and aggregates them to obtain the initial local context. The latter branch designs a learnable matrix to dynamically extract implicit local contexts for each correspondence. The output of the implicit branch is used to guide the explicit branch to adaptively expand its neighbor range. However, we observe that simply combining the representations derived from these two branches leads to sub-optimal. First, there is no interaction between the aggregated or unpooled representations from both branches, leading to a lack of mutual understanding between the nodes. It results in weak expressive and discriminative ability of network representation. Second, there are a large number of inconsistent outliers in the data, and assigning neighbors to these outliers incorrectly can easily hinder the local context learning of the network. Thus, we propose a motion injection module to address the aforementioned issues. In particular, to tackle the first problem, we introduce self-attention operations that facilitate internal interactions among nodes, enabling a thorough understanding of each other. For the second problem, we introduce motion consistency to reduce the impact of local context errors of outliers through a post-suppressing strategy. To be specific, we introduce a cross-attention operation to implement the injection of motion information, which aids in limiting the interference of outliers during the nodes' internal interactions. Additionally, to reduce the quadratic complexity limitation of the attention mechanism, we resort to the FlowAttention~\cite{wu2022flowformer} to replace traditional attention, which enhances computational efficiency while ensuring performance. Finally, the results from both branches are combined after being processed by the motion injection module.

In summary, the main contributions of this work are as follows: 1) We propose the CorrAdaptor model based on a dual-branch structure. This model adaptively adjusts the range of neighbors for each correspondence by using implicit context to guide explicit context, thereby enhancing the adaptability and stability of the model. 2)
We propose a motion injection module that introduces motion consistency to reduce the influence of local context errors from outliers on model discrimination by a post-suppression strategy. Meanwhile, we model the relationship between the neighborhood of each correspondence to obtain a more representational local context.
3) We evaluate the performance of CorrAdaptor on three distinct datasets, and our CorrAdaptor yields competitive results. In addition, our comprehensive ablation experiments showcase the advantages of our dual-branch structure and the motion injection module.

\section{Related Work}
\label{sec:related_work}

\subsection{Traditional Methods}
RANSAC~\cite{fischler1981random} and its variants \cite{raguram2012usac, barath2019magsac, torr2000mlesac, chum2005matching} are widely recognized methods for estimating geometric models through iterative sampling strategies. These approaches \cite{fischler1981random, raguram2012usac, barath2019magsac, torr2000mlesac, chum2005matching} involve selecting a minimal subset of correspondences to hypothesize a parametric model, which is verified by counting the consistent inliers. For example, USAC \cite{raguram2012usac} proposes a unified framework for incorporating multiple advancements. MAGSAC \cite{barath2019magsac} introduces $\sigma$-consensus to eliminate the need for a predefined inlier-outlier threshold, providing robustness to varying data distributions. MLESAC \cite{torr2000mlesac} extends RANSAC~\cite{fischler1981random} by maximizing the likelihood and incorporates a parameterization approach to handle nonlinear constraints. PROSAC \cite{chum2005matching} utilizes a progressive sampling strategy based on similarity rankings to achieve significant computational savings compared to RANSAC~\cite{fischler1981random}. Despite their robustness, RANSAC-based methods are sensitive to outliers \cite{jiang2021review, ma2021image, zhao2020image}, limiting their performance in scenarios with heavily contaminated initial correspondences, \ie their efficacy diminishes as outlier ratios increase \cite{ma2021image, jin2021image}. Thus, preprocessing steps like correspondence pruning are often necessary to improve their performance in outlier-rich environments \cite{yi2018learning, zhao2019nm, zhang2019learning}. 

\subsection{Learning-Based Methods}
With the advent of deep learning~\cite{wang2024gridformer,wang2022survey,yi2018learning}, numerous studies focus on leveraging learning-based methods to address the correspondence pruning task. Several pioneering works~\cite{brachmann2017dsac, ranftl2018deep, yi2018learning} utilize deep neural networks for outlier removal, yielding promising outcomes. DSAC \cite{brachmann2017dsac} introduces a differentiable variant of RANSAC~\cite{fischler1981random}, substituting its non-differentiable hypothesis selection procedure with a probabilistic approach. PointCN \cite{yi2018learning} frames the correspondence pruning task as both an inliers/outliers classification and an essential matrix regression problem. Correct matches are anticipated to be consistent in both local and global contexts, with neighbor consistency recognized as a crucial cue for distinguishing inliers from outliers. Numerous studies propose diverse strategies to mine neighbor consistency from explicit and implicit perspectives, yielding promising results. \cite{zhao2021progressive, liu2021learnable, dai2022ms2dg, liu2023progressive} utilize explicit neighborhood relation modeling, such as k-nearest neighbors, to extract consistent neighbors. LMCNet \cite{liu2021learnable} links each correspondence and its k-nearest neighbors according to their coordinate distance. CLNet \cite{zhao2021progressive} builds a KNN-based graph in feature space to obtain the local consensus score. NCMNet \cite{liu2023progressive} explores different types of k-nearest neighbors in spatial, feature, and global graph spaces, using the cross-attention mechanism to capture their relationships. In contrast to explicit methods, implicit methods enable the network to capture consistent neighbors in a learnable manner. Inspired by \cite{ying2018hierarchical}, OANet \cite{zhang2019learning} assigns correspondences to clusters in a soft assignment manner and proposes an unpooling layer to upsample the clusters for full-size prediction. Following \cite{zhang2019learning}, U-Match \cite{li2023u} builds a UNet-like multi-level architecture with pooling and unpooling operations to modulate local context. MGNet \cite{dai2024mgnet} builds local graphs from implicit and explicit aspects to leverage their complementary relationship.  However, existing learnable methods tend to construct coarse-grained local graphs, which may lead to a broad scope of local context and inconsistent correspondences. To address this, we introduce a dual-branch structure to adaptively capture the local context for each correspondence. The explicit branch is responsible for extracting a fixed range of neighbors to establish the initial local context. Subsequently, the representation derived from the implicit branch serves as guidance, facilitating the adaptive expansion of local contexts.

\section{Methodology}
\label{sec:method}

\begin{figure*}[t]
  \centering
  \includegraphics[width=1.0\textwidth]{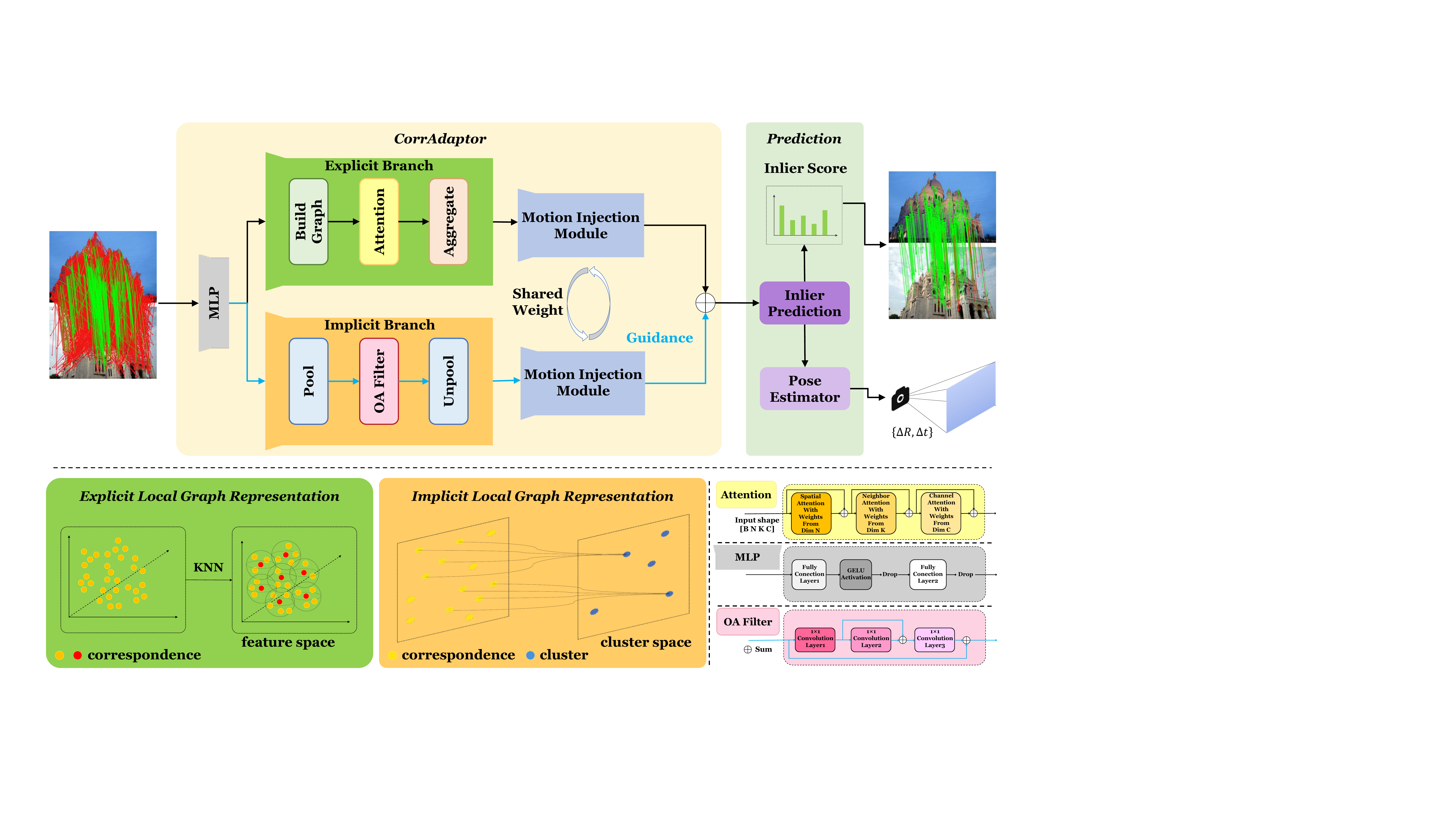}
  \caption{The architecture of our CorrAdaptor. It primarily comprises a dual-branch structure for adaptive local context contraction and a motion injection module for reducing the influence of local context errors from outliers. The explicit branch constructs a KNN-based graph for fixed-range neighborhood extraction, while the implicit branch clusters correspondences in a learnable manner.}
  \label{fig:framework—1}
\end{figure*}

\subsection{Preliminary}
Given an image pair $(I_A, I_B)$, correspondence pruning is to identify correct matches from putative correspondences and recover camera poses. Initially, feature detectors (SIFT~\cite{lowe2004distinctive}, SuperPoint~\cite{detone2018superpoint}, \etc) are used to extract feature points and descriptors. Then the initial correspondence set $C$ is formed by performing nearest neighbor matching between the extracted descriptors of feature points:
\begin{equation}
C=\{c_1, c_2, \cdots, c_N\}\in\mathbb{R}^{N\times4},
\end{equation}
where $c_i=(x_i,y_i,u_i,v_i)$ indicates the correspondence between feature point coordinates $(x_i, y_i)$ in image $I_A$ and $(u_i,v_i)$ in image $I_B$, which are normalized by camera intrinsics as in \cite{yi2018learning}.

The initial correspondence set $C$ often comprises a significant number of outliers, posing a challenge in identifying correct matches. Therefore, following \cite{zhao2021progressive}, we progressively prune $C$ into a subset of candidates $\hat{C}\in\mathbb{R}^{\hat{N}\times4}$, where $\hat{N}<N$. In addition, we obtain the inlier weight $\hat{W}$ for each candidate, which indicates the probability of each candidate as an inlier. The above process can be formulated as follows:
\begin{align}
(\hat{C}, \hat{W}) = f_{\theta}(C),
\end{align}
where $\hat{C}$ denotes the candidate set; $\hat{W}$ is the inlier weight of each candidate; $f_{\theta}$ indicates our proposed method with parameters $\theta$.

Then, we employ a weighted eight-point algorithm \cite{yi2018learning} to predict the essential matrix $\hat{E}$ for camera pose estimation. Finally, following \cite{zhao2021progressive}, we perform full-size verification to retrieve inliers that are incorrectly discarded during the pruning process. Specifically, the verification approach is conducted on the initial correspondence set $C$ to derive the predicted symmetric epipolar distance set $\hat{D}$. Note that epipolar distances with an empirical threshold ($10^{-4}$) are employed to distinguish between inliers and outliers. The above process can be formulated as follows:
\begin{equation}
\begin{aligned}
    \hat{E} = g(\hat{C}, \hat{W}), \\
    \hat{D} = h(\hat{E},C),
\end{aligned}
\end{equation}
where $g(\cdot, \cdot)$ denotes the weighted eight-point algorithm, $\hat{E}$ denotes the predicted essential matrix, $h(\cdot, \cdot)$ denotes the full-size verification, and $\hat{D}$ denotes the predicted symmetric epipolar distance set. 

\subsection{Overall Architecture}

As illustrated in Fig.~\ref{fig:framework—1}, our CorrAdaptor primarily consists of three main components: the explicit branch, the implicit branch, and the motion injection module. Initially, a dual-branch structure is employed to extract local context from explicit and implicit perspectives. The explicit branch identifies a small, fixed number of neighbors for each correspondence and aggregates their information to obtain the initial local context. Meanwhile, the implicit branch employs a learnable matrix to extract potential local context for each correspondence. 

Although explicit and implicit strategies are commonly adopted for extracting local context in recent studies~\cite{zhao2021progressive, liu2021learnable, liu2023progressive, zhang2019learning, dai2024mgnet}, we argue that outliers demonstrate inconsistency due to their random distribution, rendering the assignment of local neighbors to outliers inappropriate. In existing research, both the KNN method employed in the explicit strategy and the clustering mechanism employed in the implicit strategy force outliers to possess local neighbors. Therefore, these studies improperly integrate outliers with local contexts and complicate their differentiation from inliers.

To address this issue, we introduce the motion injection module, which leverages motion as a strong prior to rectify error arising during the learning of local context. The motion injection module can alleviate the influence of outliers and enhance the representation of inliers during local context learning, thereby effectively mitigating the impact of inappropriate local context from outliers on model discrimination.

Following the motion injection module, we guide the resultant representation from the explicit branch with the output from the implicit branch. This integration facilitates the adaptive adjustment of the local context scope for each correspondence, thereby improving local representation.
\subsection{Explicit Local Graph Representation}
In the explicit branch, we first leverage the k-nearest neighbor approach to construct the explicit local graph denoted as:
\begin{equation}
    \mathcal G_i^e = (\mathcal{V}_i, \mathcal{E}_i), i=1,2,\cdots,N,
\end{equation}
where $\mathcal{V}_i=\{f_{i1}, \cdots, f_{ik}\}$ denotes the k-nearest neighbors of $c_i$ in feature space; Following \cite{wang2019dynamic}, we construct the edge set $\mathcal{E}_i=\{e_{i1}, \cdots, e_{ik}\}$ by concatenating each correspondence feature and the residual feature with its k-nearest neighbors, which can be formulated as:
\begin{equation}
    e_{ij} = [f_i \Vert f_i-f_{ij}],
\end{equation}
where $f_i$ and $f_{ij}$ represent the feature of $c_i$ and its j-\emph{th} neighbor, and $[\cdot \Vert \cdot]$ denotes the concatenate operation along the channel dimension.

The KNN-based graph $\mathcal G^e = \{\mathcal G_1^e, \cdots, \mathcal G_N^e\}$ may not adequately model the complex relationship between correspondences. Hence, inspired by \cite{liao2024vsformer}, we employ graph attention blocks to effectively capture spatial, channel, and neighborhood relations within explicit local graphs. We elaborate the details of spatial-wise attention as below and omit the channel-wise and neighborhood-wise attention for simplicity. In the spatial attention block, we initially embed the explicit local graph $\mathcal G^e$ using the PointCN block \cite{yi2018learning}. Then, average-pooling and max-pooling operations are applied along the channel dimension to compress context. Next, element-wise summation and MLP are used to derive the spatial attention map $A_{sa}$. This attention map captures rich spatial relations between correspondences and is incorporated into the local graph. The above process can be formulated as follows:
\begin{equation}
\begin{split}
    \hat{\mathcal G^e} &= \text{PointCN}(\mathcal G^e), \\
    A_{sa} &= \text{MLP}(\text{AvgPool}(\hat{\mathcal G^e}) + \text{MaxPool}(\hat{\mathcal G^e})), \\
    \mathcal G_{sa}^e &= \hat{\mathcal G^e} \odot \sigma(A_{sa}) + \mathcal G^e, \\
\end{split}
\end{equation}

We sequentially apply spatial, neighborhood, and channel attention to extract comprehensive relations and obtain improved graph $\mathcal G_{att}^e$. Finally, followed \cite{zhao2021progressive}, we conduct neighborhood aggregation on $\mathcal G_{att}^e$ to derive the correspondence feature $F_e\in\mathbb{R}^{N\times{C}}$. The above process can be formulated as follows:
\begin{equation}
\begin{split}
    \mathcal G_{att}^e = \text{CA}(\text{NA}(\text{SA}(\mathcal G^e))), \\
    F_e = \text{AnnualConv}(\mathcal G_{att}^e). \\
\end{split}
\end{equation}
where $\text{CA}(\cdot)$, $\text{NA}(\cdot)$, $\text{SA}(\cdot)$ represent channel, neighborhood, and spatial attention block, respectively. $\text{AnnualConv}(\cdot)$ denotes the annual convolution proposed by \cite{zhao2021progressive}.

\begin{figure}[t]
    \centering
    \includegraphics[width=0.5\textwidth]{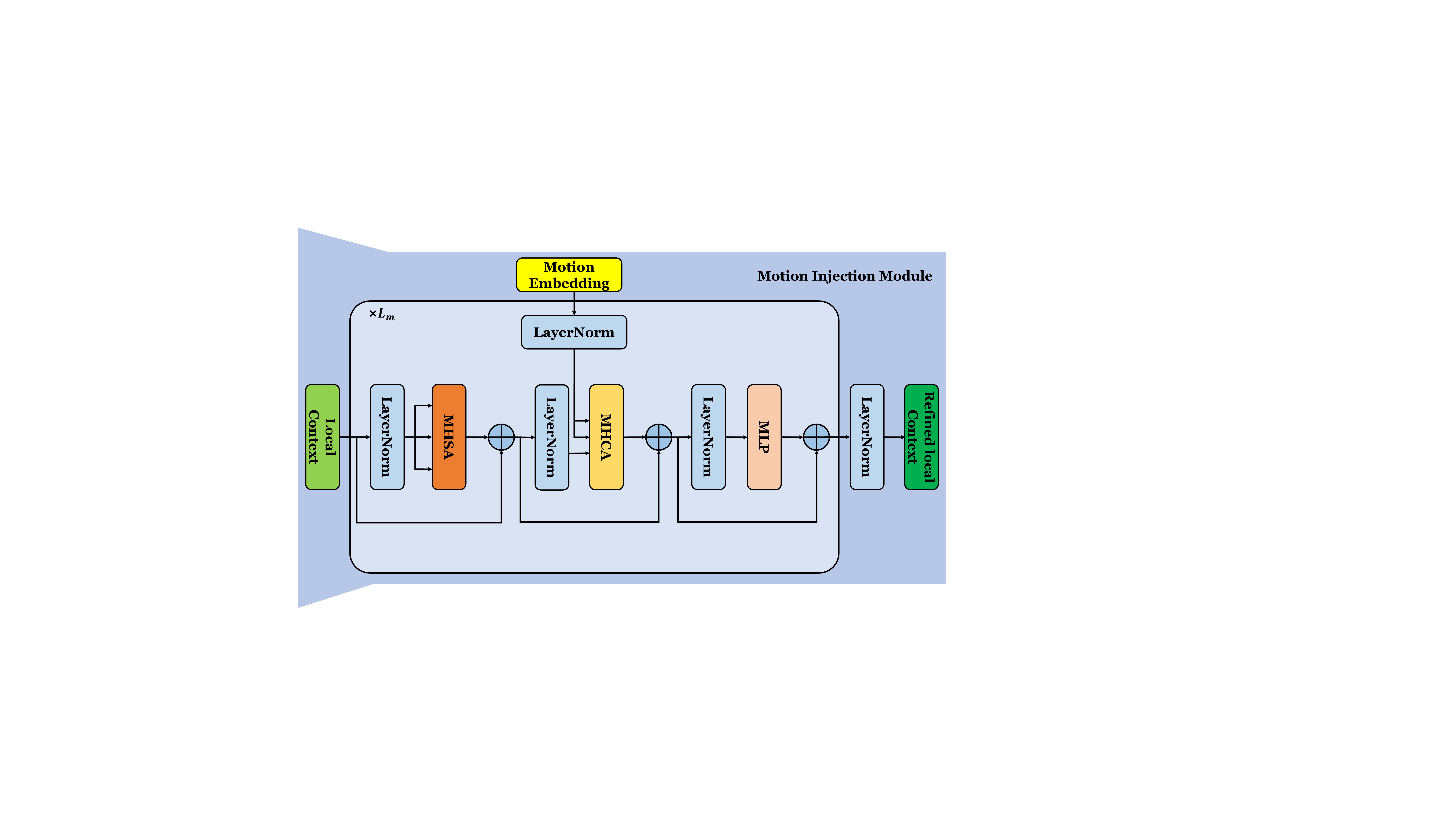}
    \vspace{-2mm}
    \caption{The architecture of the motion injection module, where MHSA is Multi-Head Self Attention and MHCA is Multi-Head Cross Attention. $L_{m}$ denotes stacking times for the motion injection module.}
    \label{fig:3}
\end{figure}

\subsection{Implicit Local Graph Representation}

The implicit branch explores the neighbor relations of correspondence in a learnable manner. In contrast to the explicit branch, it allows the network to independently determine different neighbor scopes for each correspondence. Following~\cite{zhang2019learning}, the learnable scope branch consists of three operations: the Pooling layer, the Unpooling layer, and the OA Filter block. 

Initially, the pooling layer maps correspondences into $M$ clusters via soft assignment matrix $S\in\mathbb{R}^{N\times{M}}$. Each cluster represents a weighted outcome of sparse correspondences and they form a coarse-grained graph $\mathcal{G}^i$ that captures the essential features of correspondences. Therefore, the pooling layer effectively groups neighbors consistent correspondence together in an implicit way. The above process can be formulated as follows:
\begin{equation}
    \mathcal G^i = \text{Pool}(F), \\
\end{equation}
where $F\in\mathbb{R}^{N\times{D}}$, $\mathcal G^i\in\mathbb{R}^{M\times{D}}$ represent the input correspondence features and the coarse-grained graph respectively; $D$ denotes the feature dimension, and typically $M < N$.

To enrich local representation, the OA Filter block is applied to encode complex local context on the coarsened representation graph $\mathcal{G}^i$. Subsequently, the unpooling layer restores $\mathcal{G}^i$ to its original size. The above process can be formulated as follows:
\begin{equation}
    F_i = \text{Unpool}(\text{OA}(\mathcal G^i)). \\
\end{equation}

\subsection{Motion Injection Module}
We design a motion injection module to accomplish the internal interaction of local context, ensuring that both learn higher-level features before completing the information injection from explicit to implicit. Additionally, due to the inherent impracticality of assigning neighbors to outliers, we introduce motion information into this module to reduce the interference of outliers to local context learning.

Assume there is a pair of images $A$ and $B$, each with corresponding $N$ coordinates, then the motion information can be defined as the difference in the corresponding coordinates of the two images, that is,
\begin{equation}
    \text{M}(A,B)=\{(x_i^A-x_i^B),(y_i^A-y_i^B)\}^{N}_{i=1},
\end{equation}
Motion can be seen as the collection of these $N$ vectors, each vector representing the direction from a coordinate in image $A$ towards the corresponding coordinate in image $B$. Based on this vector field, motion information can be used to capture local consistency and provide information to distinguish outliers from inliers \cite{zhang2023convmatch}.

We incorporate motion information using cross-attention \cite{vaswani2017attention}. However, upon observation, the computational efficiency of the traditional attention mechanism is significantly hindered by the typically large number of $N$, resulting in quadratic complexity. To address this issue, we insert the flow attention mechanism~\cite{wu2022flowformer} in our architecture, which is similar to CorrMAE~\cite{liao2024corrmae}. Inspired by the convergence and divergence of natural water sources, it maintains non-trivial attention while transforming the traditional $O(N^2 d)$ complexity of $(Q\times K)\times V$ into a more efficient $O(Nd^2)$ complexity of $Q\times (K\times V)$. By shifting the computational burden from sequence length to dimension, FlowAttention \cite{wu2022flowformer} achieves an efficient attention mechanism. Importantly, despite the added complexity of competition and allocation mechanisms, the performance of the output remains unaffected.

Based on the above, the structure of the motion injection module illustrated in Fig.~\ref{fig:3}, is outlined as follows:
\begin{equation}
    \begin{split}
        \hat{m}&=\text{Linear}(\text{M}),\\
        F_{e/i}&=\text{MHSA}(F_{e/i})+F_{e/i}, \\
        F_{e/i}&=\text{MHCA}(F_{e/i},~\hat{m}) +F_{e/i},  \\
        F_{e/i}&=\text{FFN}(F_{e/i})+F_{e/i}.
    \end{split}
\end{equation}
where $\text{M}$ refers to motion, MHSA denotes multi-head self-attention, and MHCA denotes multi-head cross-attention.

Firstly, motion is encoded into motion embedding using a linear projection. Then, the local context information undergoes self-attention to complete internal interactions within the context, followed by cross-attention to inject motion, where all attentions refer to the FlowAttention. Finally, the result of cross-attention is passed through an FFN to produce an output for one iteration. Each operation includes a residual connection. The entire process iterates $L_{m}$ times.

The output results of the two branches after processing a series of motion injection modules are aggregated by summing the elements, completing the guidance from the implicit branch to the explicit branch. After $L_{fusion}$ iterations, the resulting output will be used for pruning. Specifically, $F_{fusion}$ passes through a linear layer to obtain weights $w$. After sorting $w$ in reverse order, according to the pruning rate $\alpha$, the top $N\times\alpha$ correspondences are selected to complete the pruning. After $L_p$ iterations of pruning, the output is used for weight eight calculations to obtain the final result.

\subsection{Loss Function}
Following (\cite{hartley2003multiple}, \cite{ranftl2018deep}), a hybrid loss function is applied as the training objective:
\begin{equation}
\mathcal L = \mathcal L_{cls}(o, y) + \lambda \mathcal L_{reg}(\hat{E}, E),
\end{equation}
where $\mathcal L_{cls}$ denotes inliers/outliers classification loss, $\mathcal L_{reg}$ denotes the regression loss of essential matrix. $\lambda$ is a hyper-parameter to balance two losses.

Following~\cite{zhao2021progressive}, the classification loss $\mathcal L_{cls}$ is denoted as:
\begin{equation}
\mathcal L_{cls}(o, y) = \sum_{i=1}^{M} \mathcal B(\omega_i \odot o_i,y_i),
\end{equation}
where $\mathcal B(\cdot, \cdot)$ represents binary cross entropy, $o_i$ indicates the inlier weights obtained in the i-\emph{th} pruning process. $y_i$ denotes the ground-truth labels, $\omega_i$ refers to an adaptive temperature proposed by CLNet~\cite{zhao2021progressive} and $\odot$ indicates the Hadamard product.

Following~\cite{zhang2019learning}, $\mathcal L_{reg}$ represents the regression loss between the ground truth essential matrix $E$ and the predicted matrix $\hat{E}$:
\begin{equation}
\mathcal L_{reg}(\hat{E}, E) = \frac{(q^T \hat{E} p)^2}{\lVert{Ep}\rVert_{[1]}^2 + \lVert{Ep}\rVert_{[2]}^2 + \lVert{E^Tq}\rVert_{[1]}^2 + \lVert{{E^T}q}\rVert_{[2]}^2},
\end{equation}
where $p$, $q$ are the virtual coordinates generated from the ground truth essential matrix $E$, and  $\lVert{\cdot}\rVert_{[i]}$ refers to the i-\emph{th} element of the vector.

\section{Experiments}\label{sec:experiments}

\subsection{Experimental Settings}


\textbf{Implementation Details}. In the pruning strategy, two pruning blocks are utilized to reduce the putative correspondences from $N$ to $N/4$ where $\alpha$ is 0.5. For each pruning block, the number of neighbors for KNN is set to 9 and 6 for the first and second layers, respectively, and the dimension $d$ is maintained at 128. We set $L_m$ to 2, $L_p$ to 2.
Regarding training parameters, we adopt the Adam optimizer following \cite{zhang2019learning} and set the weight decay to 0. The batch size is set to 32, and a constant learning rate of 1e-3 is used throughout the training period, which spans 500k iterations.
The experiments utilize 8 V100 GPUs for computational support.
SIFT \cite{lowe2004distinctive} descriptors are utilized to generate $N=2000$ putative correspondences. Then, the keypoint coordinates are normalized using the camera intrinsics, as mentioned in \cite{yi2018learning}. We consider correspondences with small distances ($10^{-4}$) in normalized image coordinates to their ground-truth epipolar lines as true correspondences. For a pair of images, an essential matrix is estimated by RANSAC~\cite{fischler1981random} based on the predicted true correspondences, which is then decomposed into a rotation and a translation.

\begin{figure*}[t]
  \centering
  \includegraphics[width=0.85\textwidth]{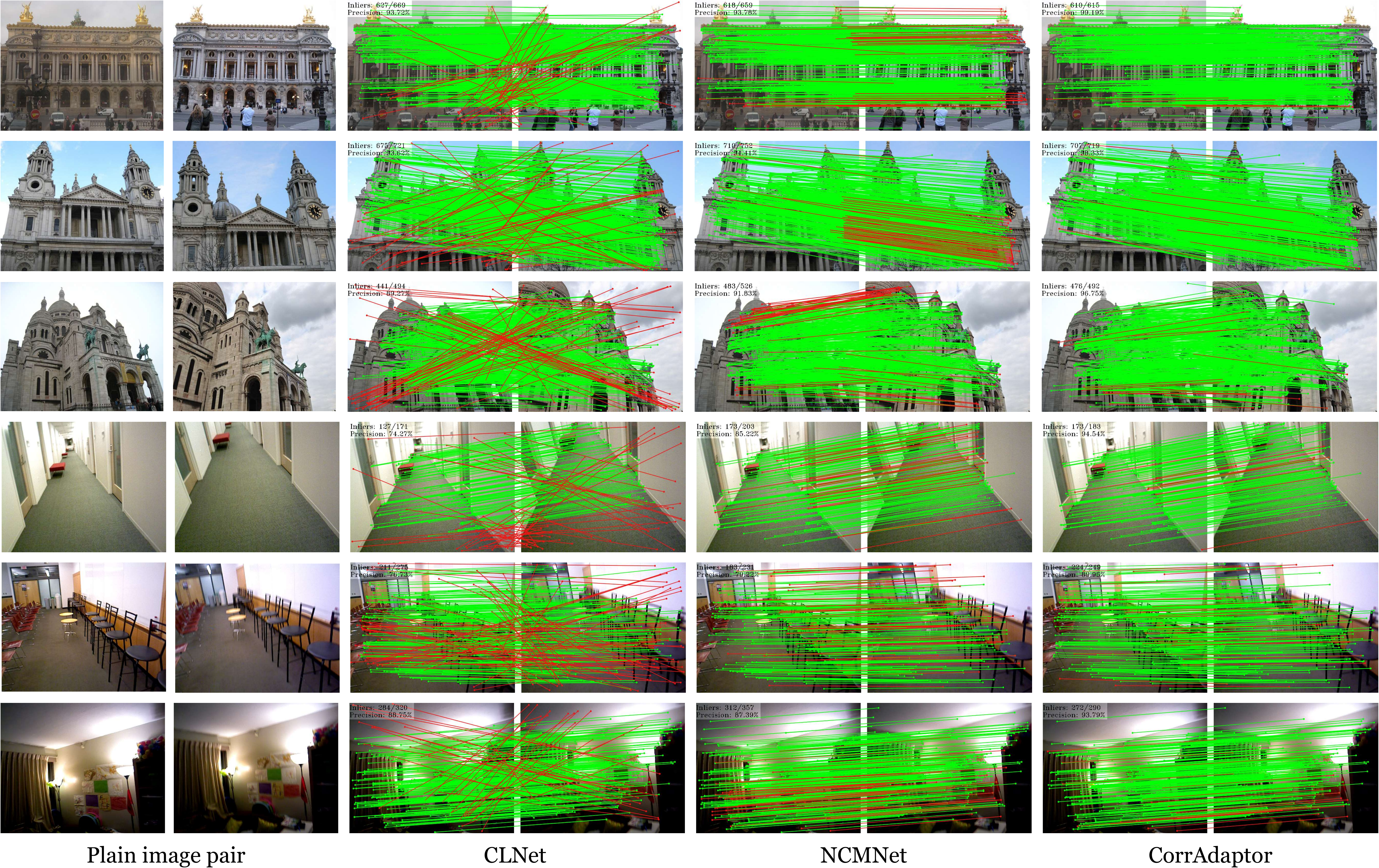}
  \caption{Visual comparison results on YFCC100M and SUN3D datasets. From left to right: the visual results of CLNet~\cite{zhao2021progressive}, NCMNet~\cite{liu2023progressive}, and the proposed CorrAdaptor. From top to bottom: the top three results on the YFCC100M dataset and the rest are results on SUN3D. In the figure,  the green denotes the true correspondence, and the red refers to the false correspondence. Our CorrAdaptor achieves the best visual results. \textbf{Zoom in for the best view}.}
  \label{fig:visual_results}

\end{figure*}

\textbf{Datasets}. We conduct experiments on the outdoor YFCC100M \cite{thomee2016yfcc100m} dataset and the indoor SUN3D \cite{xiao2013sun3d} dataset for the tasks of correspondence pruning and camera pose estimation. Meanwhile, the widely adopted HPatches \cite{balntas2017hpatches} dataset is used to evaluate the homography estimation task. YFCC100M~\cite{thomee2016yfcc100m} is used for outdoor scene analysis, which contains 68 sequences selected for training and validation purposes, and 4 sequences for testing. SUN3D \cite{xiao2013sun3d} contains 254 indoor image sequences, where 239 sequences are used for training and validation and others for testing. For YFCC100M and SUN3D, we test all methods on unknown scenes following the data division of~\cite{zhang2019learning}. HPatches \cite{balntas2017hpatches} comprises 108 sequences, with 52 sequences showcasing significant illumination changes and 56 sequences demonstrating a large variation in viewpoints.

\begin{table}[t]
\centering
\caption{Comparison results on the YFCC100M dataset~\cite{thomee2016yfcc100m}.}
\begin{tabular}{c||ccc}
\toprule
Method & Precision & Recall & F-score \\
\midrule
OANet \cite{zhang2019learning} & 68.05 & 68.41 & 68.23 \\
$\text{MS}^2\text{DG-Net}$ \cite{dai2022ms2dg} & 72.61 & 73.86 & 73.23 \\
PGFNet \cite{liu2023pgfnet} & 71.56 & 72.71 & 72.13 \\
CLNet \cite{zhao2021progressive} & 75.05 & 76.41 & 75.72 \\
ConvMatch \cite{zhang2023convmatch} & 73.12 & 74.39 & 73.75 \\
UMatch \cite{li2023u} & 73.97 & 75.72 & 74.83 \\
NCMNet \cite{liu2023progressive} & 77.24 & 78.57 & 77.90 \\
MGNet \cite{dai2024mgnet} & 76.97 & 79.35 & 78.14 \\ \midrule 
\textbf{CorrAdaptor (Ours)} & \textbf{79.23}$_{\textcolor{red}{+2.26}}$ & \textbf{80.28} $_{\textcolor{red}{+0.93}}$& \textbf{79.67}$_{\textcolor{red}{+1.53}}$ \\
\bottomrule
\end{tabular}
\label{table_1}
\end{table}

\textbf{Evaluation Metrics.} 
Camera pose estimation refers to accurately estimating the relative position relationship (\ie rotation and translation) between different camera views with identified inliers. The evaluation of error metrics is based on the angular differences between the estimated rotation/translation vectors and the corresponding ground truth vectors. For this task, we report the AUC metrics of the pose accuracy at various thresholds (5°, 10°, 20°).

Homography estimation is to find a linear image-to-image mapping in the homogeneous space. To facilitate a fair comparison across methods that yield a varying number of matches, the corner error between images is warped by the estimated homography $\hat{\mathcal{H}}$ and the ground-truth homography $\mathcal{H}$ is computed as a measure of correctness, following the approach outlined in \cite{detone2018superpoint}. Following \cite{sarlin2020superglue}, we report the AUC metrics at threshold values of 3, 5, and 10 pixels.

\subsection{Comparison Results}

\begin{table}[t]
\centering
\caption{Results of the camera pose estimation on the YFCC100M dataset~\cite{thomee2016yfcc100m}. AUC measures the results with different thresholds.}
\begin{tabular}{c||cccc}
\toprule
Method & AUC@5° & AUC@10° & AUC@20° \\
\midrule
PointCN \cite{yi2018learning} & 10.16 & 24.43 & 43.31 \\
OANet \cite{zhang2019learning} & 15.92 & 35.93 & 57.11 \\
$\text{MS}^2\text{DG-Net}$ \cite{dai2022ms2dg} & 20.61 & 42.90 & 64.26 \\
LMCNet \cite{liu2021learnable} & 22.35 & 43.57 & 63.34 \\
PGFNet \cite{liu2023pgfnet} & 24.11 & 45.97 & 65.08 \\
CLNet \cite{zhao2021progressive} & 25.28 & 45.82 & 65.44 \\
ConvMatch \cite{zhang2023convmatch} & 26.83 & 49.14 & 67.91 \\
UMatch \cite{li2023u} & 30.84 & 52.04 & 69.65 \\
NCMNet \cite{liu2023progressive} & 34.51 & 55.34 & 72.40 \\
MGNet \cite{dai2024mgnet} & 32.32 & 53.40 & 71.59 \\ \midrule 
\textbf{CorrAdaptor (Ours)} & \textbf{41.02}$_{\textcolor{red}{+6.51}}$ & \textbf{61.67} $_{\textcolor{red}{+6.33}}$& \textbf{76.98}$_{\textcolor{red}{+4.58}}$ \\
\bottomrule
\end{tabular}
\label{table_2}
\end{table}

\textbf{Correspondence Pruning.} The comparison results on the YFCC100M \cite{thomee2016yfcc100m} are listed in Table~\ref{table_1}. As shown, our CorrAdaptor outperforms the state-of-the-art models across all three metrics. To be specific, compared with the recent NCMNet~\cite{liu2023progressive}, our CorrAdaptor achieves a performance improvement of $2.26$ in Precision. Among the comparison methods, the recent MGNet~\cite{dai2024mgnet} is the most similar to our method, which also considers constructing the local graph from implicit and explicit aspects. Compared with MGNet~\cite{dai2024mgnet}, our CorrAdaptor exhibits performance enhancements of 2.26, 0.93, and 1.53 in Precision, Recall, and F-score, respectively. This further verifies the superior performance of our model.

\begin{table}[t]
\centering
\caption{Results of the camera pose estimation on the Sun3d dataset~\cite{xiao2013sun3d}, where AUC measures the values with different thresholds.}
\begin{tabular}{c||cccc}
\toprule
Method & AUC@5° & AUC@10° & AUC@20° \\
\midrule
PointCN \cite{yi2018learning} & 3.05 & 10.00 & 24.06 \\
OANet \cite{zhang2019learning} & 5.93 & 16.91 & 34.32 \\
$\text{MS}^2\text{DG-Net}$ \cite{dai2022ms2dg} & 5.88 & 16.83 & 34.28 \\
LMCNet \cite{liu2021learnable} & 7.08 & 19.09 & 37.15 \\
ConvMatch \cite{zhang2023convmatch} & 8.76 & 22.23 & 40.49 \\ \midrule 
\textbf{CorrAdaptor (Ours)} & \textbf{9.33}$_{\textcolor{red}{+0.57}}$ & \textbf{22.66}$_{\textcolor{red}{+0.43}}$ & \textbf{40.85}$_{\textcolor{red}{+0.36}}$ \\
\bottomrule
\end{tabular}
\label{table_3}
\end{table}
\textbf{Camera Pose Estimation.} We also analyse the performance comparison for the camera pose estimation task on both outdoor and indoor scenes. Table~\ref{table_2} displays the results on the outdoor dataset YFCC100M for the camera pose estimation task. CorrAdaptor outperforms the previous best results by 6.51, 6.33, and 4.58 for the AUC metrics at 5°, 10°, and 20°, respectively, showcasing the strengths of our approach. 

Furthermore, we evaluate CorrAdaptor's performance on the indoor dataset Sun3d \cite{xiao2013sun3d}, which is presented in Table~\ref{table_3}. The Sun3d \cite{xiao2013sun3d} dataset encompasses a multitude of frames that exhibit poor texture, repetitive elements, and instances of self-occlusion, thus making it inherently more challenging compared to YFCC100M \cite{thomee2016yfcc100m}. Despite these complexities, CorrAdaptor surpasses the performance of previous methods, indicating its effectiveness in handling challenging indoor scenes. Additionally, we visually compare recent representational methods (\ie CLNet~\cite{zhao2021progressive}, NCMNet~\cite{liu2023progressive}). The visual results are shown in Fig.~\ref{fig:visual_results}, where our CorrAdaptor obtains the best performance under
various challenging scenes.

\begin{table}[t]
\centering
\caption{Quantitative comparison results of the homography estimation on HPatches, where the results are measured by AUC with different thresholds.}
\begin{tabular}{c||cccc}
\toprule
Method & AUC@3° & AUC@5° & AUC@10° \\
\midrule
CLNet \cite{zhao2021progressive} & 33.38 & 40.90 & 50.00 \\
OANet \cite{zhang2019learning} & 54.79 & 64.79 & 75.45 \\
$\text{MS}^2\text{DG-Net}$ \cite{dai2022ms2dg} & 57.24 & 67.80 & 78.64 \\
UMatch \cite{li2023u} & 57.13 & 67.40 & 77.67 \\
MGNet \cite{dai2024mgnet} & 60.42 & 71.80 & 82.76 \\ \midrule 
\textbf{CorrAdaptor (Ours)} & \textbf{61.53}$_{\textcolor{red}{+1.11}}$ & \textbf{72.72}$_{\textcolor{red}{+0.9}}$ & \textbf{83.26}$_{\textcolor{red}{+0.5}}$ \\
\bottomrule
\end{tabular}
\label{table_4}
\end{table}

\textbf{Homography Estimation.} To further evaluate the performance of the proposed method, we compare our method with recent methods for the task of homography estimation. The comparison results are shown in Table~\ref{table_4}. We can find that our CorrAdaptor has the best performance and outperforms others by a large margin. Specifically, it outperforms the previous best method by approximately 1.11, 0.9, and 0.5 in AUC@3°, AUC@5°, and AUC@10°, respectively. These improvements highlight the effectiveness of CorrAdaptor in accurately estimating homography.

\begin{table}[t]
\centering
\caption{Ablation study for the structures of CorrAdaptor.}
\begin{tabular}{c||ccc}
\toprule
Model& AUC@5° & AUC@10° & AUC@20° \\
\midrule
single branch-implicit & 35.53 & 55.53 & 71.74 \\
single branch-explicit & 37.94 & 58.40 & 74.51 \\
\textbf{dual branch (CorrAdaptor)} & \textbf{41.02} & \textbf{61.67} & \textbf{76.98} \\
\bottomrule
\end{tabular}
\label{table_5}
\end{table}

\begin{table}[t]
\centering
\caption{Ablation study for the numbers of clusters in the implicit branch.}
\begin{tabular}{c||cccc}
\toprule
Model & Params. &AUC@5° & AUC@10° & AUC@20° \\
\midrule
500  & 9.09M& 40.22 & 60.69 & 76.39 \\
1000  & 18.62M&39.17 & 59.91 & 75.88 \\
\textbf{250 (CorrAdaptor)}  &\textbf{6.57M} & \textbf{41.02} & \textbf{61.67} & \textbf{76.98} \\
\bottomrule
\end{tabular}
\label{table_6}
\end{table}

\begin{table}[!htbp]
\centering
\caption{Ablation study of FlowAttention used in our model. IPS means Iteration Per Second.}
\resizebox{0.98\linewidth}{!}{\begin{tabular}{c||ccccc}
\toprule
Model & AUC@5° & AUC@10° & AUC@20° & IPS \\
\midrule
Plain Attention \cite{vaswani2017attention} & 39.53 & 60.10 & 76.04 & 1.2 \\
\textbf{FlowAttention~(CorrAdaptor)} & \textbf{41.02} & \textbf{61.67} & \textbf{76.98} & \textbf{1.5} \\
\bottomrule
\end{tabular}}
\label{table_7}
\end{table}

\begin{table}[!htbp]
\centering
\caption{Ablation study of motion prior in the motion injection module.}
\begin{tabular}{c||ccccc}
\toprule
Model & AUC@5° & AUC@10° & AUC@20°  \\
\midrule
w/o. motion & 38.26 & 58.18 & 74.62 & \\
\textbf{w. motion~(CorrAdaptor)} & \textbf{41.02} & \textbf{61.67} & \textbf{76.98} & \\
\bottomrule
\end{tabular}
\label{table_8}
\end{table}

\begin{table}[!htbp]
\centering
\caption{Ablation study of stacking times for motion injection module.}
\begin{tabular}{l||cccccc}
\toprule
Stack Times & Params. &AUC@5° & AUC@10° & AUC@20°\\
\midrule
1 & 4.45M&37.04 & 58.29 & 74.88\\
4 & 10.82M&40.26 & 61.18 & 76.89\\
\textbf{2 (CorrAdaptor)} & \textbf{6.57M}&\textbf{41.02} & \textbf{61.67} & \textbf{76.98}\\
\bottomrule
\end{tabular}
\label{table_9}

\end{table}

\subsection{Ablation Study}

We conduct a series of ablation experiments on the YFCC100M dataset to verify the effectiveness of each module in our method

\textbf{Single branch vs. dual branch.} As shown in Table~\ref{table_5}, we explore the dual-branch structure of CorrAdaptor. It can be seen that our CorrAdaptor with a two-branch structure significantly outperforms models with explicit branches or implicit single branches. Specifically, our CorrAdaptor achieves a performance improvement of 15.4\% over the model with the single implicit branch. In addition, the model with the explicit branch outperforms the model with the implicit branch (see the results in Row-1 and Row-2). The experimental results are consistent with our motivation to use explicit branches in the model to obtain the initial local context. Thus, the ablation results show that the unique dual-branch structure of CorrAdaptor is reasonable.

\textbf{The number of clusters in the implicit branch.}
As shown in Table \ref{table_6}, we investigate the number of clusters in the implicit branches of CorrAdaptor, which directly affects the model's local context learning ability. We set different numbers of clusters (\ie 250, 500, 1000) in the models. The results show that our CorrAdaptor with 250 clusters can effectively capture local context for correspondence pruning. In addition, our CorrAdaptor contains a small number of parameters, which shows that our CorrAdaptor achieves a better trade-off between performance and complexity.

\textbf{FlowAttention in our CorrAdaptor.} We also conduct the ablation study to verify the effectiveness of the FlowAttention used in our CorrAdaptor. To be specific, we replace the FlowAttention by using the plain Attention~\cite{vaswani2017attention} in the motion injection module to perform experiments. The results are shown in Table~\ref{table_7}. Compared with the model using the plain Attention~\cite{vaswani2017attention}, our  CorrAdaptor with FlowAttention obtains performance improvements of 1.49 AUC@5°, 1.57 AUC@10°, and 0.94 AUC@20°, respectively. This is mainly because Flowattention introduces a competition mechanism in attention, which can suppress the representation of outliers in the local context. In addition, the Flowattention used in our model can improve inference efficiency (see Col-5).

\textbf{Motion prior.}
Table~\ref{table_8} serves to validate the effectiveness of leveraging motion as prior information to mitigate the impact of inappropriate local context from outliers. In the absence of motion within the motion injection module, we simply integrate representation derived from the dual-branch structure, which leads to a noticeable decline in performance. This underscores the crucial role that motion information plays in mitigating the impact of outliers during the learning process of the local context.

\textbf{Stacking times for motion injection module.} 
Table~\ref{table_9} shows the results of varying the number of stacking layers in the motion injection module. The results show that stacking each branch twice is the most effective configuration. It achieves a better trade-off between performance and complexity.

\section{Conclusion}
\label{sec:conclusion}
In this paper, we introduce CorrAdaptor, a novel architecture designed to improve pixel-level correspondences in computer vision and robotics tasks. Our approach adaptively adjusts local contexts by employing a dual-branch structure that combines explicit and implicit local graph learning, enhancing model robustness and adaptability to complex image variations. Additionally, the motion injection module is integrated into the network to integrate motion consistency to suppress the impact of outliers and refine local context learning, leading to substantial performance improvements. Extensive results show that CorrAdaptor achieves state-of-the-art performance across various correspondence-based tasks. 

\section*{Acknowledgement}
This work was supported in part by China Mobile Zijin Innovation Insititute (No. NR2310J7M), and the National
Natural Science Foundation of China (GrantNo. 62372223).  

\clearpage
\bibliography{mybibfile}

\begin{thebibliography}{42}
\providecommand{\natexlab}[1]{#1}
\providecommand{\url}[1]{\texttt{#1}}
\expandafter\ifx\csname urlstyle\endcsname\relax
  \providecommand{\doi}[1]{doi: #1}\else
  \providecommand{\doi}{doi: \begingroup \urlstyle{rm}\Url}\fi

\bibitem[Balntas et~al.(2017)Balntas, Lenc, Vedaldi, and Mikolajczyk]{balntas2017hpatches}
V.~Balntas, K.~Lenc, A.~Vedaldi, and K.~Mikolajczyk.
\newblock Hpatches: A benchmark and evaluation of handcrafted and learned local descriptors.
\newblock In \emph{Proceedings of the IEEE Conference on Computer Vision and Pattern Recognition}, pages 5173--5182, 2017.

\bibitem[Barath et~al.(2019)Barath, Matas, and Noskova]{barath2019magsac}
D.~Barath, J.~Matas, and J.~Noskova.
\newblock Magsac: marginalizing sample consensus.
\newblock In \emph{Proceedings of the IEEE Conference on Computer Vision and Pattern Recognition}, pages 10197--10205, 2019.

\bibitem[Brachmann et~al.(2017)Brachmann, Krull, Nowozin, Shotton, Michel, Gumhold, and Rother]{brachmann2017dsac}
E.~Brachmann, A.~Krull, S.~Nowozin, J.~Shotton, F.~Michel, S.~Gumhold, and C.~Rother.
\newblock Dsac-differentiable ransac for camera localization.
\newblock In \emph{Proceedings of the IEEE Conference on Computer Vision and Pattern Recognition}, pages 6684--6692, 2017.

\bibitem[Chum and Matas(2005)]{chum2005matching}
O.~Chum and J.~Matas.
\newblock Matching with prosac-progressive sample consensus.
\newblock In \emph{Proceedings of the IEEE Computer Society Conference on Computer Vision and Pattern Recognition}, pages 220--226, 2005.

\bibitem[Dai et~al.(2022)Dai, Liu, Ma, Wei, Lai, Yang, and Chen]{dai2022ms2dg}
L.~Dai, Y.~Liu, J.~Ma, L.~Wei, T.~Lai, C.~Yang, and R.~Chen.
\newblock Ms2dg-net: Progressive correspondence learning via multiple sparse semantics dynamic graph.
\newblock In \emph{Proceedings of the IEEE Conference on Computer Vision and Pattern Recognition}, pages 8973--8982, 2022.

\bibitem[Dai et~al.(2024)Dai, Du, Zhang, and Tang]{dai2024mgnet}
L.~Dai, X.~Du, H.~Zhang, and J.~Tang.
\newblock Mgnet: Learning correspondences via multiple graphs.
\newblock \emph{arXiv preprint arXiv:2401.04984}, 2024.

\bibitem[DeTone et~al.(2018)DeTone, Malisiewicz, and Rabinovich]{detone2018superpoint}
D.~DeTone, T.~Malisiewicz, and A.~Rabinovich.
\newblock Superpoint: Self-supervised interest point detection and description.
\newblock In \emph{Proceedings of the IEEE Conference on Computer Vision and Pattern Recognition Workshops}, pages 224--236, 2018.

\bibitem[Fischler and Bolles(1981)]{fischler1981random}
M.~A. Fischler and R.~C. Bolles.
\newblock Random sample consensus: a paradigm for model fitting with applications to image analysis and automated cartography.
\newblock \emph{Communications of the ACM}, 24\penalty0 (6):\penalty0 381--395, 1981.

\bibitem[Hartley and Zisserman(2003)]{hartley2003multiple}
R.~Hartley and A.~Zisserman.
\newblock \emph{Multiple view geometry in computer vision}.
\newblock Cambridge university press, 2003.

\bibitem[Jiang et~al.(2021)Jiang, Ma, Xiao, Shao, and Guo]{jiang2021review}
X.~Jiang, J.~Ma, G.~Xiao, Z.~Shao, and X.~Guo.
\newblock A review of multimodal image matching: Methods and applications.
\newblock \emph{Information Fusion}, 73:\penalty0 22--71, 2021.

\bibitem[Jin et~al.(2021)Jin, Mishkin, Mishchuk, Matas, Fua, Yi, and Trulls]{jin2021image}
Y.~Jin, D.~Mishkin, A.~Mishchuk, J.~Matas, P.~Fua, K.~M. Yi, and E.~Trulls.
\newblock Image matching across wide baselines: From paper to practice.
\newblock \emph{International Journal of Computer Vision}, 129\penalty0 (2):\penalty0 517--547, 2021.

\bibitem[Li et~al.(2023)Li, Zhang, and Ma]{li2023u}
Z.~Li, S.~Zhang, and J.~Ma.
\newblock U-match: two-view correspondence learning with hierarchy-aware local context aggregation.
\newblock In \emph{Proceedings of the International Joint Conference on Artificial Intelligence}, pages 1169--1176, 2023.

\bibitem[Liao et~al.(2024{\natexlab{a}})Liao, Zhang, Xiao, Li, Wang, and Ye]{liao2024corrmae}
T.~Liao, X.~Zhang, G.~Xiao, M.~Li, T.~Wang, and M.~Ye.
\newblock Corrmae: Pre-training correspondence transformers with masked autoencoder.
\newblock \emph{arXiv preprint arXiv:2406.05773}, 2024{\natexlab{a}}.

\bibitem[Liao et~al.(2024{\natexlab{b}})Liao, Zhang, Zhao, Wang, and Xiao]{liao2024vsformer}
T.~Liao, X.~Zhang, L.~Zhao, T.~Wang, and G.~Xiao.
\newblock Vsformer: Visual-spatial fusion transformer for correspondence pruning.
\newblock In \emph{Proceedings of the AAAI Conference on Artificial Intelligence}, pages 3369--3377, 2024{\natexlab{b}}.

\bibitem[Liu and Yang(2023)]{liu2023progressive}
X.~Liu and J.~Yang.
\newblock Progressive neighbor consistency mining for correspondence pruning.
\newblock In \emph{Proceedings of the IEEE Conference on Computer Vision and Pattern Recognition}, pages 9527--9537, 2023.

\bibitem[Liu et~al.(2023)Liu, Xiao, Chen, and Ma]{liu2023pgfnet}
X.~Liu, G.~Xiao, R.~Chen, and J.~Ma.
\newblock Pgfnet: Preference-guided filtering network for two-view correspondence learning.
\newblock \emph{IEEE Transactions on Image Processing}, 32:\penalty0 1367--1378, 2023.

\bibitem[Liu et~al.(2021)Liu, Liu, Lin, Dong, and Wang]{liu2021learnable}
Y.~Liu, L.~Liu, C.~Lin, Z.~Dong, and W.~Wang.
\newblock Learnable motion coherence for correspondence pruning.
\newblock In \emph{Proceedings of the IEEE Conference on Computer Vision and Pattern Recognition}, pages 3237--3246, 2021.

\bibitem[Lowe(2004)]{lowe2004distinctive}
D.~G. Lowe.
\newblock Distinctive image features from scale-invariant keypoints.
\newblock \emph{International Journal of Computer Vision}, 60:\penalty0 91--110, 2004.

\bibitem[Ma et~al.(2019)Ma, Ma, and Li]{ma2019infrared}
J.~Ma, Y.~Ma, and C.~Li.
\newblock Infrared and visible image fusion methods and applications: A survey.
\newblock \emph{Information Fusion}, 45:\penalty0 153--178, 2019.

\bibitem[Ma et~al.(2021)Ma, Jiang, Fan, Jiang, and Yan]{ma2021image}
J.~Ma, X.~Jiang, A.~Fan, J.~Jiang, and J.~Yan.
\newblock Image matching from handcrafted to deep features: A survey.
\newblock \emph{International Journal of Computer Vision}, 129\penalty0 (1):\penalty0 23--79, 2021.

\bibitem[Mur-Artal et~al.(2015)Mur-Artal, Montiel, and Tardos]{mur2015orb}
R.~Mur-Artal, J.~M.~M. Montiel, and J.~D. Tardos.
\newblock Orb-slam: a versatile and accurate monocular slam system.
\newblock \emph{IEEE Transactions on Robotics}, 31\penalty0 (5):\penalty0 1147--1163, 2015.

\bibitem[Raguram et~al.(2012)Raguram, Chum, Pollefeys, Matas, and Frahm]{raguram2012usac}
R.~Raguram, O.~Chum, M.~Pollefeys, J.~Matas, and J.-M. Frahm.
\newblock Usac: A universal framework for random sample consensus.
\newblock \emph{IEEE Transactions on Pattern Analysis and Machine Intelligence}, 35\penalty0 (8):\penalty0 2022--2038, 2012.

\bibitem[Ranftl and Koltun(2018)]{ranftl2018deep}
R.~Ranftl and V.~Koltun.
\newblock Deep fundamental matrix estimation.
\newblock In \emph{Proceedings of the European Conference on Computer Vision}, pages 284--299, 2018.

\bibitem[Sarlin et~al.(2020)Sarlin, DeTone, Malisiewicz, and Rabinovich]{sarlin2020superglue}
P.-E. Sarlin, D.~DeTone, T.~Malisiewicz, and A.~Rabinovich.
\newblock Superglue: Learning feature matching with graph neural networks.
\newblock In \emph{Proceedings of the IEEE Conference on Computer Vision and Pattern Recognition}, pages 4938--4947, 2020.

\bibitem[Sattler et~al.(2018)Sattler, Maddern, Toft, Torii, Hammarstrand, Stenborg, Safari, Okutomi, Pollefeys, Sivic, et~al.]{sattler2018benchmarking}
T.~Sattler, W.~Maddern, C.~Toft, A.~Torii, L.~Hammarstrand, E.~Stenborg, D.~Safari, M.~Okutomi, M.~Pollefeys, J.~Sivic, et~al.
\newblock Benchmarking 6dof outdoor visual localization in changing conditions.
\newblock In \emph{Proceedings of the IEEE Conference on Computer Vision and Pattern Recognition}, pages 8601--8610, 2018.

\bibitem[Schonberger and Frahm(2016)]{schonberger2016structure}
J.~L. Schonberger and J.-M. Frahm.
\newblock Structure-from-motion revisited.
\newblock In \emph{Proceedings of the IEEE Conference on Computer Vision and Pattern Recognition}, pages 4104--4113, 2016.

\bibitem[Thomee et~al.(2016)Thomee, Shamma, Friedland, Elizalde, Ni, Poland, Borth, and Li]{thomee2016yfcc100m}
B.~Thomee, D.~A. Shamma, G.~Friedland, B.~Elizalde, K.~Ni, D.~Poland, D.~Borth, and L.-J. Li.
\newblock Yfcc100m: The new data in multimedia research.
\newblock \emph{Communications of the ACM}, 59\penalty0 (2):\penalty0 64--73, 2016.

\bibitem[Torr and Zisserman(2000)]{torr2000mlesac}
P.~H. Torr and A.~Zisserman.
\newblock Mlesac: A new robust estimator with application to estimating image geometry.
\newblock \emph{Computer Vision and Image Understanding}, 78\penalty0 (1):\penalty0 138--156, 2000.

\bibitem[Vaswani et~al.(2017)Vaswani, Shazeer, Parmar, Uszkoreit, Jones, Gomez, Kaiser, and Polosukhin]{vaswani2017attention}
A.~Vaswani, N.~Shazeer, N.~Parmar, J.~Uszkoreit, L.~Jones, A.~N. Gomez, {\L}.~Kaiser, and I.~Polosukhin.
\newblock Attention is all you need.
\newblock In \emph{Proceedings of Advances in Neural Information Processing Systems}, 2017.

\bibitem[Wang et~al.(2022)Wang, Zhang, Chen, Luo, Deng, Lu, Cao, Liu, Li, and Zafeiriou]{wang2022survey}
T.~Wang, K.~Zhang, X.~Chen, W.~Luo, J.~Deng, T.~Lu, X.~Cao, W.~Liu, H.~Li, and S.~Zafeiriou.
\newblock A survey of deep face restoration: Denoise, super-resolution, deblur, artifact removal.
\newblock \emph{arXiv preprint arXiv:2211.02831}, 2022.

\bibitem[Wang et~al.(2024)Wang, Zhang, Shao, Luo, Stenger, Lu, Kim, Liu, and Li]{wang2024gridformer}
T.~Wang, K.~Zhang, Z.~Shao, W.~Luo, B.~Stenger, T.~Lu, T.-K. Kim, W.~Liu, and H.~Li.
\newblock Gridformer: Residual dense transformer with grid structure for image restoration in adverse weather conditions.
\newblock \emph{International Journal of Computer Vision}, pages 1--23, 2024.

\bibitem[Wang et~al.(2019)Wang, Sun, Liu, Sarma, Bronstein, and Solomon]{wang2019dynamic}
Y.~Wang, Y.~Sun, Z.~Liu, S.~E. Sarma, M.~M. Bronstein, and J.~M. Solomon.
\newblock Dynamic graph cnn for learning on point clouds.
\newblock \emph{ACM Transactions on Graphics}, 38\penalty0 (5):\penalty0 1--12, 2019.

\bibitem[Wu et~al.(2022)Wu, Wu, Xu, Wang, and Long]{wu2022flowformer}
H.~Wu, J.~Wu, J.~Xu, J.~Wang, and M.~Long.
\newblock Flowformer: Linearizing transformers with conservation flows.
\newblock \emph{arXiv preprint arXiv:2202.06258}, 2022.

\bibitem[Xiao et~al.(2013)Xiao, Owens, and Torralba]{xiao2013sun3d}
J.~Xiao, A.~Owens, and A.~Torralba.
\newblock Sun3d: A database of big spaces reconstructed using sfm and object labels.
\newblock In \emph{Proceedings of the IEEE International Conference on Computer Vision}, pages 1625--1632, 2013.

\bibitem[Yi et~al.(2018)Yi, Trulls, Ono, Lepetit, Salzmann, and Fua]{yi2018learning}
K.~M. Yi, E.~Trulls, Y.~Ono, V.~Lepetit, M.~Salzmann, and P.~Fua.
\newblock Learning to find good correspondences.
\newblock In \emph{Proceedings of the IEEE Conference on Computer Vision and Pattern Recognition}, pages 2666--2674, 2018.

\bibitem[Ying et~al.(2018)Ying, You, Morris, Ren, Hamilton, and Leskovec]{ying2018hierarchical}
Z.~Ying, J.~You, C.~Morris, X.~Ren, W.~Hamilton, and J.~Leskovec.
\newblock Hierarchical graph representation learning with differentiable pooling.
\newblock In \emph{Proceedings of Advances in Neural Information Processing Systems}, volume~31, 2018.

\bibitem[Zhang et~al.(2019)Zhang, Sun, Luo, Yao, Zhou, Shen, Chen, Quan, and Liao]{zhang2019learning}
J.~Zhang, D.~Sun, Z.~Luo, A.~Yao, L.~Zhou, T.~Shen, Y.~Chen, L.~Quan, and H.~Liao.
\newblock Learning two-view correspondences and geometry using order-aware network.
\newblock In \emph{Proceedings of the IEEE International Conference on Computer Vision}, pages 5845--5854, 2019.

\bibitem[Zhang and Ma(2023)]{zhang2023convmatch}
S.~Zhang and J.~Ma.
\newblock Convmatch: Rethinking network design for two-view correspondence learning.
\newblock \emph{IEEE Transactions on Pattern Analysis and Machine Intelligence}, 2023.

\bibitem[Zhao et~al.(2019)Zhao, Cao, Li, Li, and Yang]{zhao2019nm}
C.~Zhao, Z.~Cao, C.~Li, X.~Li, and J.~Yang.
\newblock Nm-net: Mining reliable neighbors for robust feature correspondences.
\newblock In \emph{Proceedings of the IEEE Conference on Computer Vision and Pattern Recognition}, pages 215--224, 2019.

\bibitem[Zhao et~al.(2020)Zhao, Cao, Yang, Xian, and Li]{zhao2020image}
C.~Zhao, Z.~Cao, J.~Yang, K.~Xian, and X.~Li.
\newblock Image feature correspondence selection: A comparative study and a new contribution.
\newblock \emph{IEEE Transactions on Image Processing}, 29:\penalty0 3506--3519, 2020.

\bibitem[Zhao et~al.(2021)Zhao, Ge, Zhu, Zhao, Li, and Salzmann]{zhao2021progressive}
C.~Zhao, Y.~Ge, F.~Zhu, R.~Zhao, H.~Li, and M.~Salzmann.
\newblock Progressive correspondence pruning by consensus learning.
\newblock In \emph{Proceedings of the IEEE International Conference on Computer Vision}, pages 6464--6473, 2021.

\bibitem[Zhong et~al.(2021)Zhong, Xiao, Zheng, Lu, and Ma]{zhong2021t}
Z.~Zhong, G.~Xiao, L.~Zheng, Y.~Lu, and J.~Ma.
\newblock T-net: Effective permutation-equivariant network for two-view correspondence learning.
\newblock In \emph{Proceedings of the IEEE International Conference on Computer Vision}, pages 1950--1959, 2021.

\end{thebibliography}

\end{document}